%% file: main.tex
\documentclass{article}

\usepackage[preprint, nonatbib]{neurips_2020_ml4ad}

\usepackage[utf8]{inputenc} 
\usepackage[T1]{fontenc}    
\usepackage{hyperref}       
\usepackage{url}            
\usepackage{booktabs}       
\usepackage{amsfonts}       
\usepackage{nicefrac}       
\usepackage{microtype}      
\usepackage{graphicx}
\usepackage{subcaption}
\usepackage[dvipsnames]{xcolor}

\title{A Customizable Dynamic Scenario Modeling and Data Generation Platform for Autonomous Driving}

\author{%
  Jay Shenoy \\
  UC Berkeley \\
  \texttt{jayshenoy@berkeley.edu} \\
  \And
  Edward Kim \\
  UC Berkeley \\
  \texttt{ek65@eecs.berkeley.edu} \\
  \And
  Xiangyu Yue \\
  UC Berkeley \\
  \texttt{xyyue@berkeley.edu} \\
  \And
  Taesung Park \\
  UC Berkeley \\
  \texttt{taesung\_park@berkeley.edu} \\
  \And
  Daniel Fremont \\
  UC Santa Cruz \\
  \texttt{dfremont@ucsc.edu} \\
  \And
  Alberto Sangiovanni-Vincentelli \\
  UC Berkeley \\
  \texttt{alberto@berkeley.edu} \\
  \And
  Sanjit Seshia \\
  UC Berkeley \\
  \texttt{sseshia@eecs.berkeley.edu} \\
}

\input{macros}

\begin{document}

\maketitle

\begin{abstract}
Safely interacting with humans is a significant challenge for autonomous driving. The performance of this interaction depends on machine learning-based modules of an autopilot, such as perception, behavior prediction, and planning. These modules require training datasets with high-quality labels and a diverse range of realistic dynamic behaviors. Consequently, training such modules to handle rare scenarios is difficult because they are, by definition, rarely represented in real-world datasets. Hence, there is a practical need to augment datasets with synthetic data covering these rare scenarios. In this paper, we present a platform to \emph{model} dynamic and interactive scenarios, \emph{generate} the scenarios in simulation with different modalities of labeled sensor data, and \emph{collect} this information for data augmentation. To our knowledge, this is the first integrated platform for these tasks specialized to the autonomous driving domain.
\end{abstract}

\section{Introduction}
\input{introduction}

\section{Related Work}
\input{related_work}

\section{\scenic: A Formal Dynamic Scenario Description Language}
\input{dynamic_scenic}
\label{sec:scenic}

\section{Platform Pipeline}
\input{platform_pipeline}

\section{Experiments}
\input{experiments}



\section{Conclusion}
\input{conclusion}


\bibliographystyle{ieeetr}
\bibliography{references}

\end{document}

%% file: macros.tex
\newcommand{\scenic}{\textsc{Scenic}}

%% file: introduction.tex
Dynamic interactions with humans is a significant challenge for autonomous driving. Perception, behavior prediction, and planning modules of autonomous vehicle software stack play crucial roles in interacting with humans on the road. To train these machine learning-based modules, high quality labeled sensor data with diverse and realistic dynamic behaviors is essential. Handling rare scenarios for these modules is especially difficult because of the, by definition, small proportion of such scenarios in real-world training datasets. Hence, there is a practical need to augment these datasets with synthetic data. However, in simulation, modeling dynamic and interactive environments with complex spatiotemporal relations among multiple agents is expensive, and is a serious barrier to scalable data augmentation. 

We present a platform to model and generate dynamic scenarios, and collect labeled sensor data with different sensor modalities.
To our knowledge, this is the first integrated platform for these tasks specialized to the autonomous driving domain.
We build on the \scenic{} scenario description language for modeling and generating scenarios~\cite{static_scenic}, extended to dynamic scenarios in \cite{dynamic_scenic}.
\scenic{} does not provide flexible features for generating sensor data, and in the context of data augmentation has only been used to generate static 2-dimensional RGB images~\cite{static_scenic,fremont-cav20}.
In this work, we add features to \scenic{} enabling it to serve as a dynamic scenario data generation platform, with the ability to: (1) configure arbitrary number and types of sensors on the ego vehicle in simulation as they would be mounted on a real autonomous car, (2) record data from these sensors along with corresponding ground-truth labels, (3) extract specific types of sensor data from a recorded simulation, and (4) visualize each type of recorded sensor data and labels. We open-source our platform~\cite{platform}, along with 340GB of labeled sensor data generated from example driving scenarios encoded in \scenic{}.

\begin{figure}
  \centering
  \includegraphics[width=\linewidth]{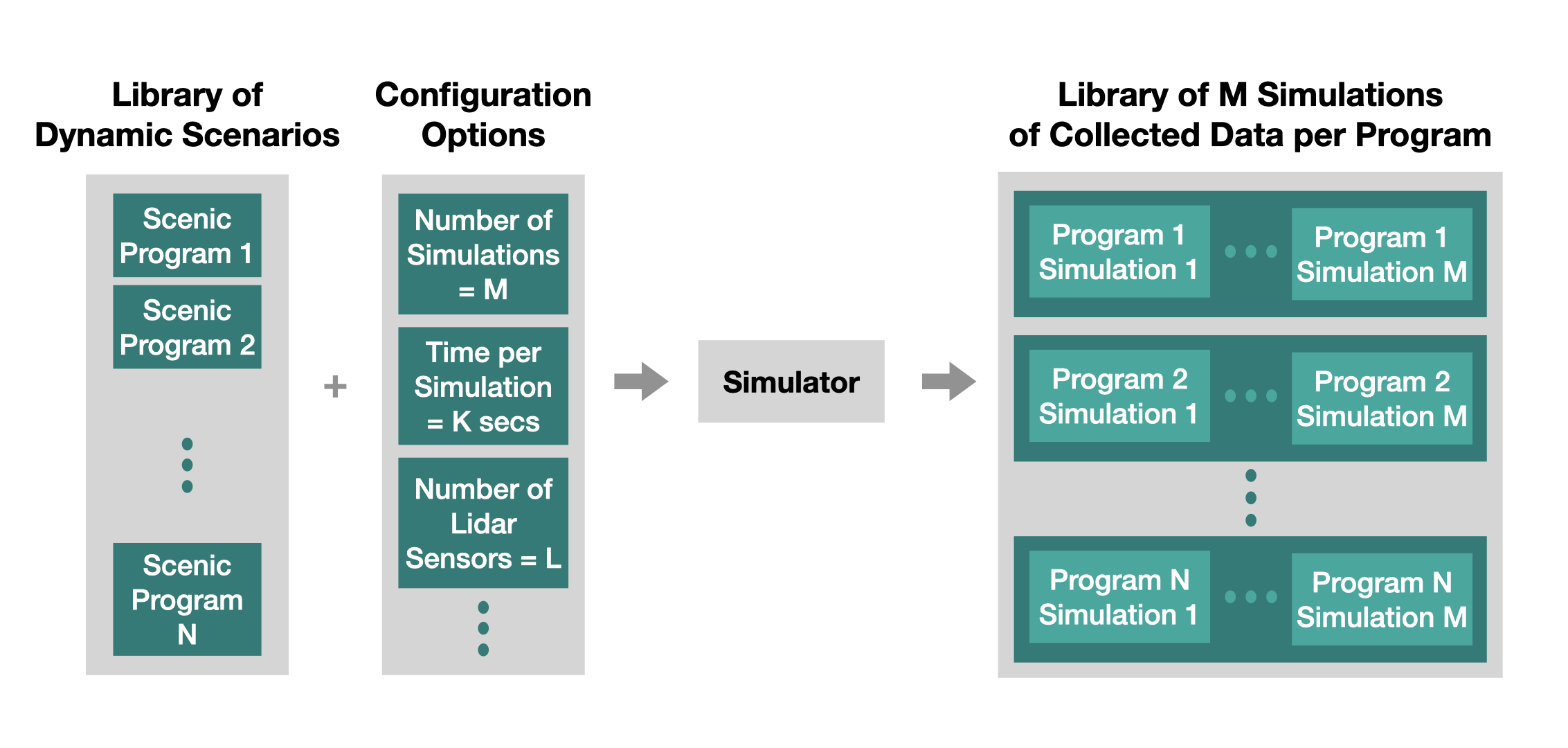}
  \caption{Overview of our platform's pipeline.}
  \label{fig:platform_flow}
\end{figure}


%% file: related_work.tex
Several learning-based methods for scene generation have been proposed. Hierarchical spatial priors between furniture are learned in~\cite{wang2019planit, yu2011make}, which are then integrated into hand-crafted cost functions for generating layouts of indoor scenes. In~\cite{eslami2016attend}, a generative model is proposed to sequentially add objects into scenes. In~\cite{kar2019meta}, a dataset generator is parametrized with a neural network. Generative Adversarial Networks (GANs)~\cite{goodfellow2014generative} have also used for data augmentation ~\cite{liang2017recurrent,marchesi2017megapixel}. In contrast to such methods, \scenic{} does not require training nor a pre-existing dataset to construct generative models. Also, as a probabilistic programming language, \scenic{} can easily incorporate declarative constraints, while producing data in an \emph{explainable}, programmatic fashion requiring only a simulator.

Recently, domain-specific scenario description languages have been proposed for autonomous cars. The Paracosm language~\cite{paracosm} models dynamic scenarios with a reactive and synchronous model of computation. Unlike \scenic{}, it lacks probability distributions and declarative constraints, as well as constructs like \scenic{}'s interrupts which enable easy customization of generic dynamic behavior models. The Measurable Scenario Description Language (M-SDL)~\cite{msdl} provides similar constructs as \scenic{}, but with some distinguishing features: (i) \scenic{} provides a higher-level, declarative way to specify geometric constraints, and (ii) \scenic{} is fundamentally a probabilistic programming language, as opposed to M-SDL, whose distributions are optional.

%% file: dynamic_scenic.tex
\subsection{Scenario Modelling}
\scenic~\cite{static_scenic,dynamic_scenic} is a formal scenario description language developed to model static and dynamic, multi-agent scenarios to help designing and testing autonomous systems such as self-driving cars. More precisely, \scenic{} is a probabilistic programming language that enables users to specify distributions over scenes (i.e., configurations of objects in space as well as their features) and dynamic behaviors. Furthermore, users can construct objects in an intuitive imperative style while simultaneously imposing declarative hard (i.e., strict) and soft (i.e., probabilistic) constraints. \scenic{}'s concise, readable syntax simplifies modelling \emph{spatial} and \emph{temporal} relationships. This syntax provides building blocks to encode typical geometric relations occurring in real-world driving scenarios, which would otherwise require complex non-linear expressions and constraints, as well as temporal constructs such as interrupts to model complex dynamic and reactive behaviors in a modular fashion. 

\subsection{Scenario Generation}
A \scenic{} program not only serves as a scenario description but also as a source of synthetic data. An execution of the \scenic{} program samples an environment according to the probabilistic distributions defined in the program. At the beginning of program execution, \scenic{} samples the static aspects of the environment, e.g. the initial positions and orientations of objects, to define the initial condition for a simulation. As the simulation runs, \scenic{} may sample further random variables pertaining to the dynamic aspects of the environment, for example delays between events. The \scenic{} program and an interfaced simulator communicate every simulation timestep to generate reactive behaviors. The simulator sends all objects' ground-truth states (e.g., speed, heading, and position) to a server running the program. Given this information, the server chooses and sends to the client (i.e., the simulator) the next action (e.g. set throttle, steering angle, brake, etc) to be simulated for each agent during the next simulation timestep. 

\begin{figure}
  \centering
  \includegraphics[width=0.9\linewidth]{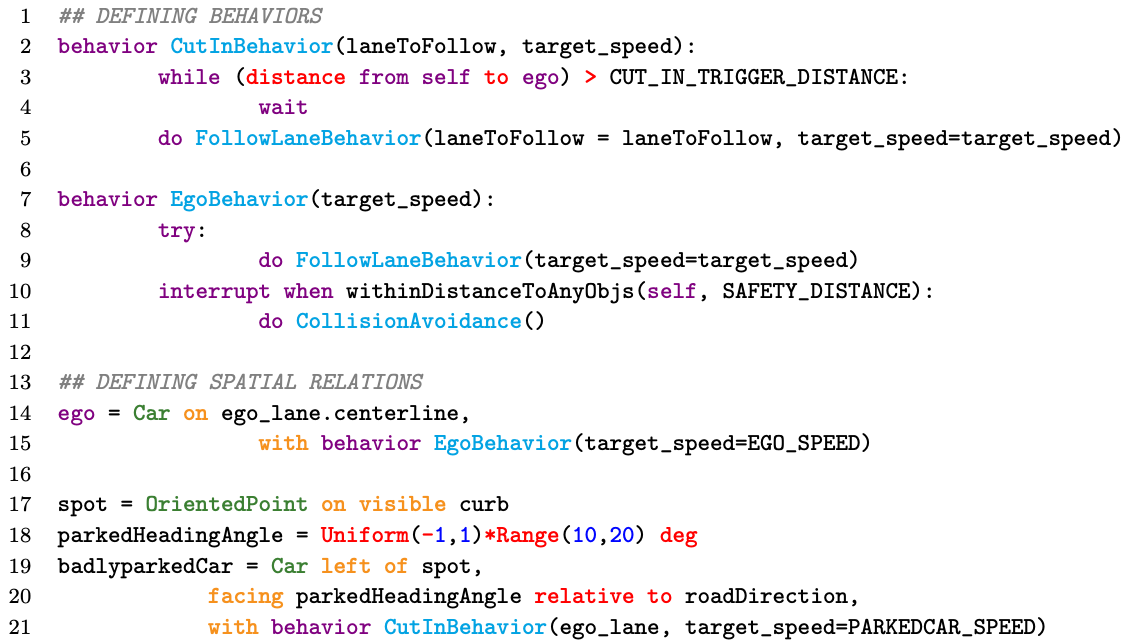}
  \caption{Part of a \scenic{} program modeling a badly-parked car pulling into the ego's driving lane.}
  \label{fig:badlyparked}
\end{figure}

For example, consider a scenario where a badly-parked car by the curb suddenly pulls into the lane that the ego vehicle is driving on. A snippet of a \scenic{} program modeling this scenario is shown in Figure~\ref{fig:badlyparked}. Data generated with this program is shown in Figure~\ref{fig:annotations}.

%% file: platform_pipeline.tex
The main contribution of this paper is a platform to generate any size of synthetic dataset covering dynamic scenarios of interest. We used the CARLA simulator~\cite{carla} as a simulation and data generation engine for this platform. However, \scenic{} is simulator-independent and can be interfaced with different simulators to collect data if the they support relevant functionalities. 

As illustrated in Figure \ref{fig:platform_flow}, the input to the platform is a batch of \scenic{} programs, each modeling a particular dynamic scenario. In addition, one can configure any number of camera and lidar sensors, each with its own set of parameters (e.g., sensor position, resolution and field of view for video cameras, or rotational frequency for lidar). The duration of each simulation as well as the number of simulations to run per scenario are adjustable. The \scenic{} programs and configuration settings are compiled and sent to CARLA to output a dataset of recorded scenarios, complete with ground-truth annotations for each sensor. These are RGB video and lidar data with ground-truth annotations for semantic segmentation, depth information, labeled point clouds, and 2D/3D bounding boxes as shown in Figure \ref{fig:annotations}. The details of how to use our platform are available at~\cite{platform}.

Our implementation also comes with an API to browse the synthetic datasets output by the platform, offering a way of programmatically interacting with annotated data in order to train and evaluate machine learning models. For instance, the API includes helper functions to project ground-truth 3D bounding box annotations to 2D bounding boxes on any given camera (implicitly computing the camera's calibration matrix), and other functions to visualize semantic point clouds with color labels.

%% file: experiments.tex
\begin{figure}
\centering
\begin{subfigure}[b]{.45\linewidth}
\includegraphics[width=\linewidth]{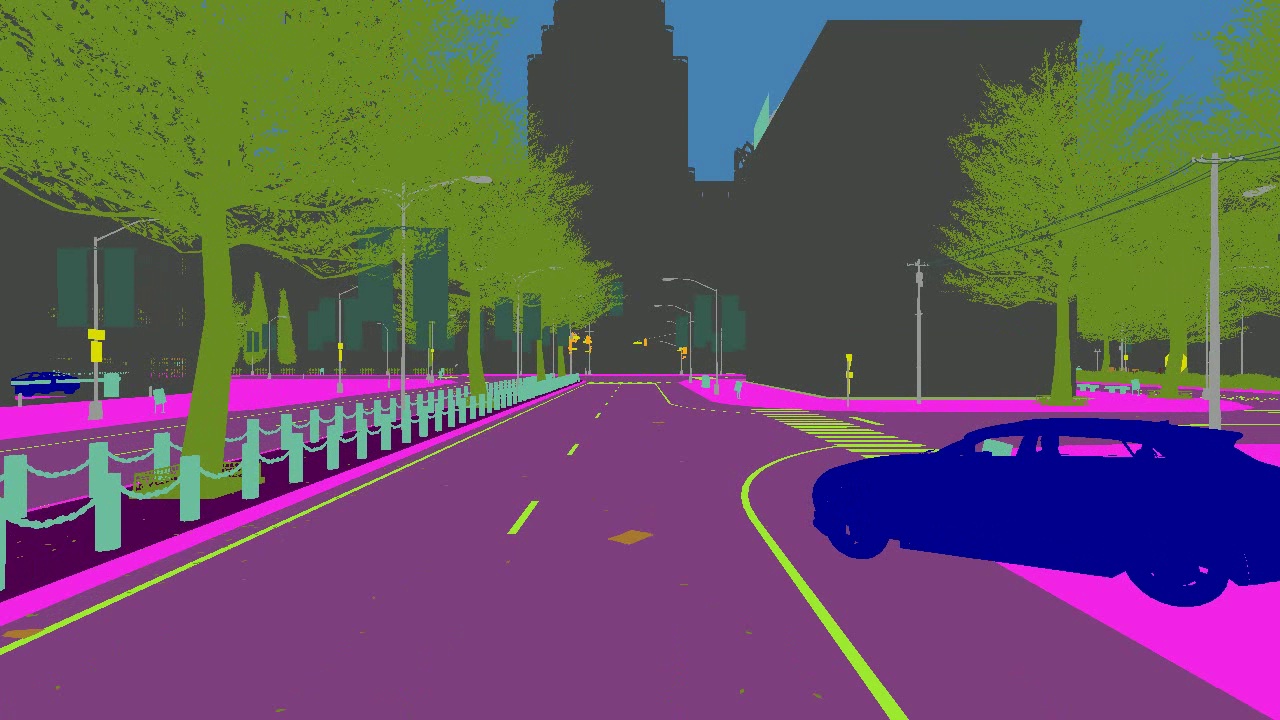}
\caption{Semantic Segmentation}
\label{fig:semantic}
\end{subfigure}
\begin{subfigure}[b]{.45\linewidth}
\includegraphics[width=\linewidth]{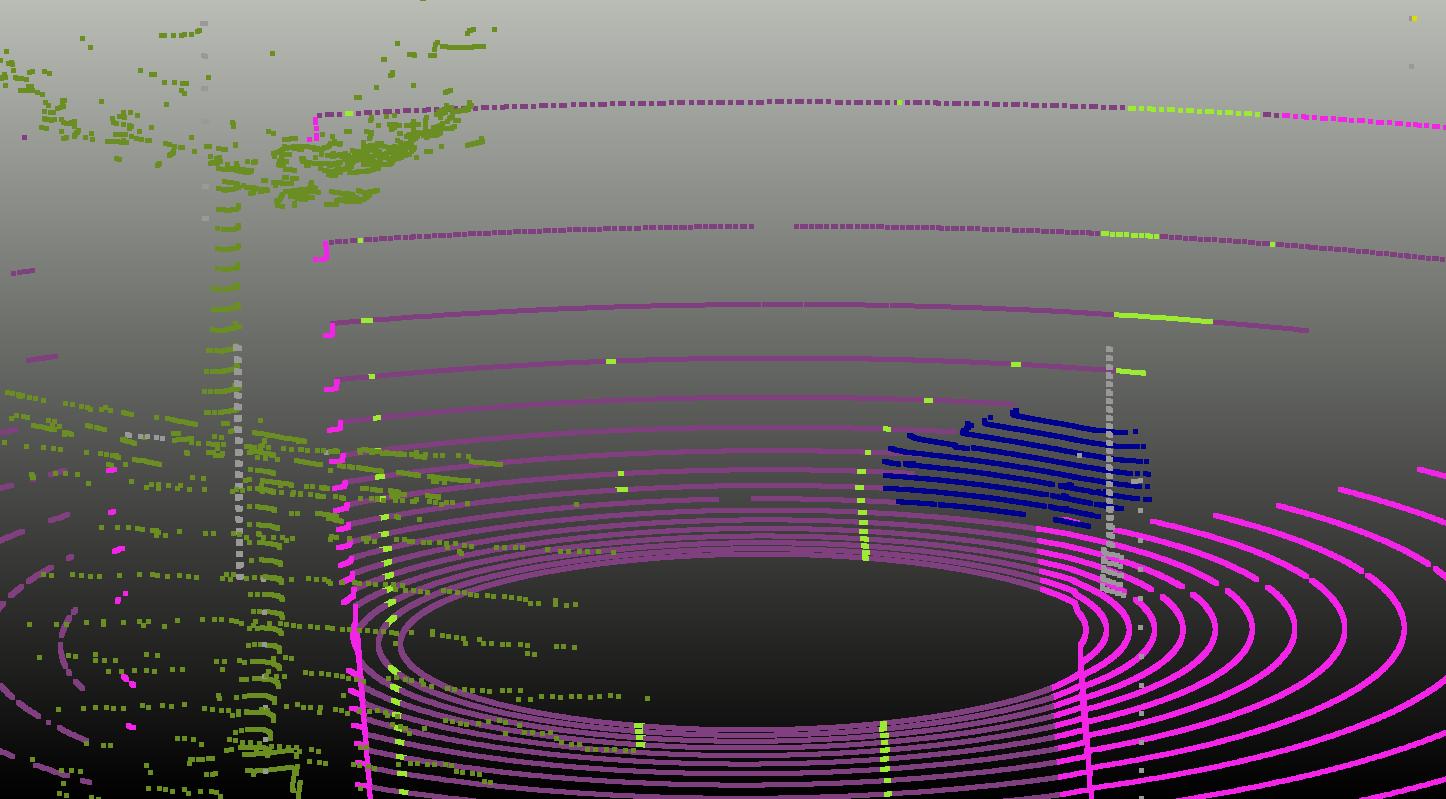}
\caption{Lidar with Semantic Labels}
\label{fig:lidar}
\end{subfigure}

\begin{subfigure}[b]{.30\linewidth}
\includegraphics[width=\linewidth]{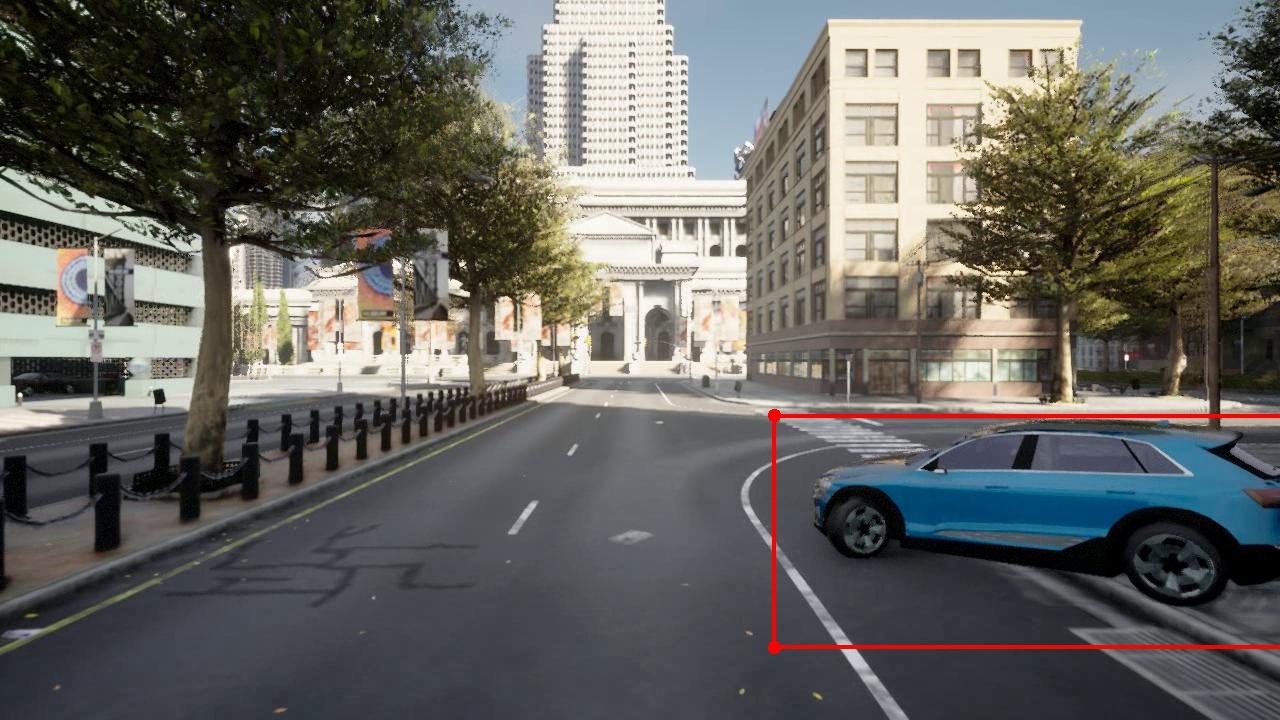}
\caption{2D Bounding Boxes}
\label{fig:bbox2d}
\end{subfigure}
\begin{subfigure}[b]{.30\linewidth}
\includegraphics[width=\linewidth]{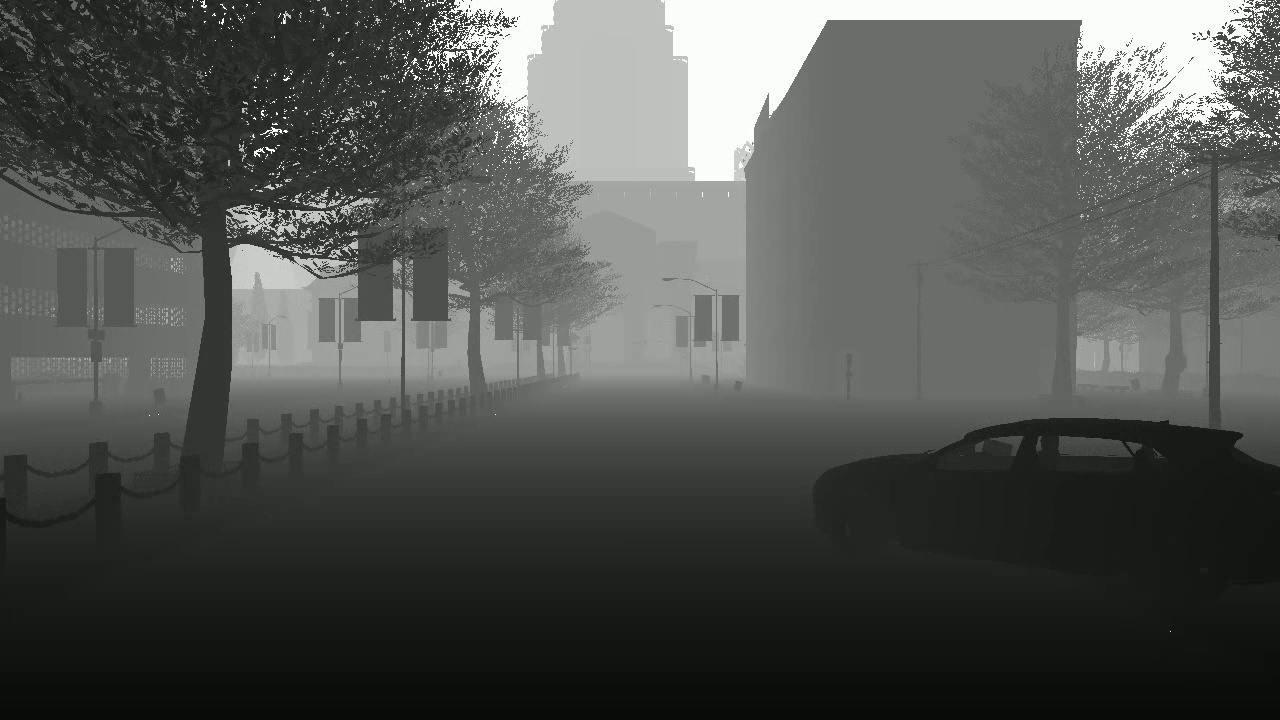}
\caption{Depth Image}
\label{fig:depth}
\end{subfigure}
\begin{subfigure}[b]{.30\linewidth}
\includegraphics[width=\linewidth]{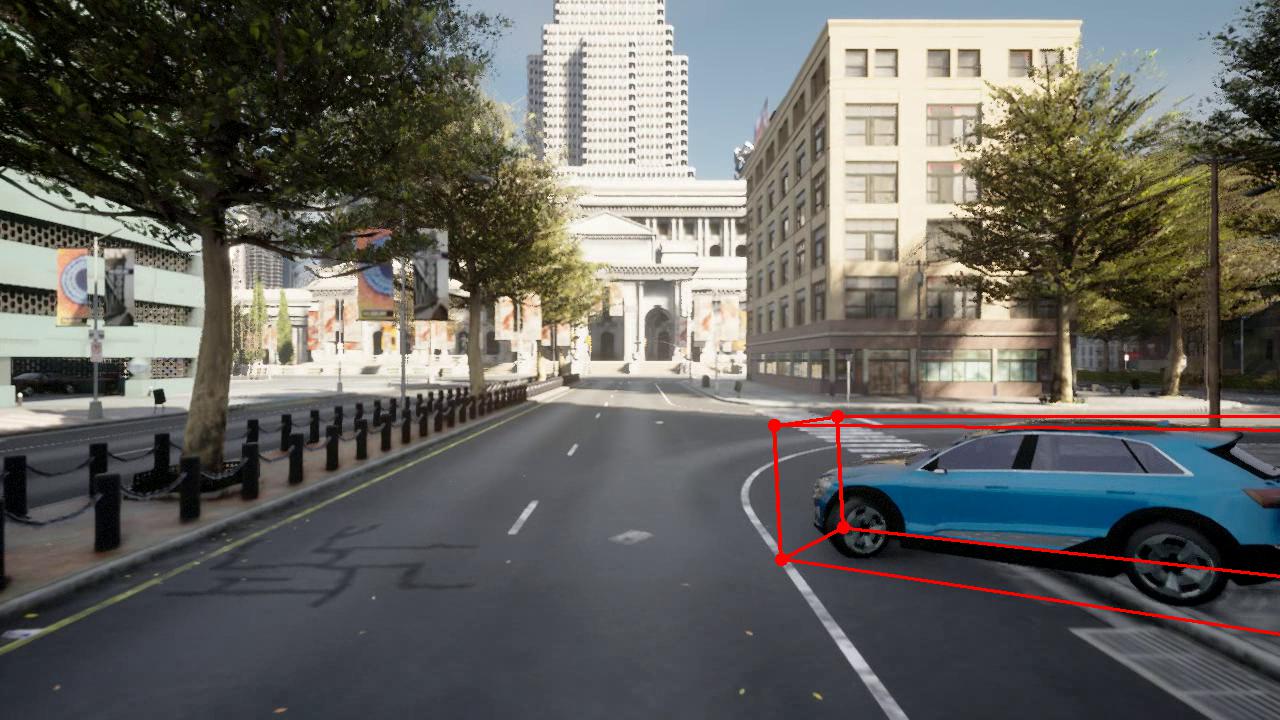}
\caption{3D Bounding Boxes}
\label{fig:bbox3d}
\end{subfigure}

\caption{Collected Sensor Data and Labels using the \scenic{} program described in Figure~\ref{fig:badlyparked}}
\label{fig:annotations}
\end{figure}

To demonstrate the effectiveness of the platform, we created a synthetic dataset of hundreds of simulated video recordings. This data were generated from a variety of \scenic{} programs representing a range of different driving scenarios. Many of these scenarios were adapted from the CARLA Autonomous Driving Challenge Scenarios \cite{carlachallenge}, which are derived from a pre-crash typology reported by the National Highway Traffic Safety Administration \cite{nhtsa}. Others were chosen from the Voyage's Open Autonomous Safety \cite{voyage}, an open-source library containing commonly-encountered road scenarios used for AV safety testing. Additional scenarios capturing high-density traffic situations, such as bumper-to-bumper lines of cars and slow-moving intersectional left turns, were also included. Each scenario was implemented as a \scenic{} program and  small variations such as background vehicles were added to the scenarios provided in \cite{carlachallenge} and \cite{voyage} in order to make the simulations more realistic.

For the synthetic dataset, an RGB camera was used to record front-facing videos for the ego vehicle, along with ground-truth depth and semantic segmentation data. In addition, a lidar sensor recorded 360-degree sweeps of the scene, including semantic labels for each point. The simulator directly reported 3D bounding boxes for objects in the scene, whose coordinates were transformed to be relative to the ego. Figure \ref{fig:annotations} illustrates the provided annotations on a single frame. The video camera and lidar sensor were placed at the same location on the ego vehicle's hood, exactly 2.4 meters off the ground. The video camera recorded 15 frames per second at a resolution of 1280x720, with a 90 degree horizontal field of view (FoV). Per-pixel-depth information and semantic labels were also captured. The lidar sensor ran at a frequency of approximately 15 sweeps per second, recording about 22,000 points per sweep. 
A complete description of all the semantic tags and exact sensor configurations can be found at \cite{platform}. 

%% file: conclusion.tex
We presented a platform to model dynamic and interactive scenarios of interest, generate such scenarios in simulation, collect annotated sensor data, and visualize the recorded data. Our platform allows researchers and developers to configure sensors as installed on their real autonomous car to help them reduce the domain gap between the real world and the synthetic dataset. We hope that our platform can assist in designing systems to safely interact with humans on the road by allowing fine-grained control over the distributions of different types of scenarios in the dataset.


%% file: main.bbl
\begin{thebibliography}{10}

\bibitem{static_scenic}
D.~Fremont, X.~Yue, T.~Dreossi, S.~Ghosh, A.~L. Sangiovanni-Vincentelli, and
  S.~A. Seshia, ``Scenic: A language for scenario specification and scene
  generation,'' {\em Programming Language Design and Implementation (PLDI)},
  2018.

\bibitem{dynamic_scenic}
D.~J. Fremont, E.~Kim, T.~Dreossi, S.~Ghosh, X.~Yue, A.~L.
  Sangiovanni-Vincentelli, and S.~A. Seshia, ``Scenic: A language for scenario
  specification and data generation.'' \url{http://arxiv.org/abs/2010.06580},
  2020.

\bibitem{fremont-cav20}
D.~J. Fremont, J.~Chiu, D.~D. Margineantu, D.~Osipychev, and S.~A. Seshia,
  ``Formal analysis and redesign of a neural network-based aircraft taxiing
  system with {VerifAI},'' in {\em 32nd International Conference on Computer
  Aided Verification (CAV)}, July 2020.

\bibitem{platform}
``Platform code base.'' \url{https://github.com/jayshenoy/scenic}.

\bibitem{wang2019planit}
K.~Wang, Y.-A. Lin, B.~Weissmann, M.~Savva, A.~X. Chang, and D.~Ritchie,
  ``Planit: Planning and instantiating indoor scenes with relation graph and
  spatial prior networks,'' {\em ACM Transactions on Graphics (TOG)}, vol.~38,
  no.~4, pp.~1--15, 2019.

\bibitem{yu2011make}
L.~F. Yu, S.~K. Yeung, C.~K. Tang, D.~Terzopoulos, T.~F. Chan, and S.~J. Osher,
  ``Make it home: automatic optimization of furniture arrangement,'' {\em ACM
  Transactions on Graphics (TOG)-Proceedings of ACM SIGGRAPH 2011, v. 30,(4),
  July 2011, article no. 86}, vol.~30, no.~4, 2011.

\bibitem{eslami2016attend}
S.~A. Eslami, N.~Heess, T.~Weber, Y.~Tassa, D.~Szepesvari, G.~E. Hinton, {\em
  et~al.}, ``Attend, infer, repeat: Fast scene understanding with generative
  models,'' in {\em Advances in Neural Information Processing Systems},
  pp.~3225--3233, 2016.

\bibitem{kar2019meta}
A.~Kar, A.~Prakash, M.-Y. Liu, E.~Cameracci, J.~Yuan, M.~Rusiniak, D.~Acuna,
  A.~Torralba, and S.~Fidler, ``Meta-sim: Learning to generate synthetic
  datasets,'' in {\em Proceedings of the IEEE International Conference on
  Computer Vision}, pp.~4551--4560, 2019.

\bibitem{goodfellow2014generative}
I.~Goodfellow, J.~Pouget-Abadie, M.~Mirza, B.~Xu, D.~Warde-Farley, S.~Ozair,
  A.~Courville, and Y.~Bengio, ``Generative adversarial nets,'' in {\em
  Advances in neural information processing systems}, pp.~2672--2680, 2014.

\bibitem{liang2017recurrent}
X.~Liang, Z.~Hu, H.~Zhang, C.~Gan, and E.~P. Xing, ``Recurrent topic-transition
  gan for visual paragraph generation,'' {\em arXiv preprint arXiv:1703.07022},
  2017.

\bibitem{marchesi2017megapixel}
M.~Marchesi, ``Megapixel size image creation using generative adversarial
  networks,'' {\em arXiv preprint arXiv:1706.00082}, 2017.

\bibitem{paracosm}
R.~Majumdar, A.~Mathur, M.~Pirron, L.~Stegner, and D.~Zufferey, ``Paracosm: A
  language and tool for testing autonomous driving systems,'' 2019.

\bibitem{msdl}
Foretellix, ``Measurable scenario description language.''
  \url{https://www.foretellix.com/wp-content/uploads/2020/07/M-SDL_LRM_OS.pdf},
  2020.

\bibitem{carla}
A.~Dosovitskiy, G.~Ros, F.~Codevilla, A.~M. L{\'{o}}pez, and V.~Koltun,
  ``{CARLA:} an open urban driving simulator,'' {\em CoRR},
  vol.~abs/1711.03938, 2017.

\bibitem{carlachallenge}
``Carla autonomous driving challenge.'' \url{https://carlachallenge.org/}.

\bibitem{nhtsa}
W.~G. Najm, J.~D. Smith, and M.~Yanagisawa, ``Pre-crash scenario typology for
  crash avoidance research,'' 2007.

\bibitem{voyage}
``Voyage open autonomous safety.'' \url{https://oas.voyage.auto/}.

\end{thebibliography}
